\documentclass[letterpaper, 10 pt, conference]{ieeeconf}  

\IEEEoverridecommandlockouts                              

\usepackage{mathtools}
\usepackage{amsmath}  
\usepackage{amssymb}  
\usepackage{bm}       
\usepackage{graphicx} 
\usepackage{hyperref} 
\usepackage{cite}     
\usepackage{url}      
\usepackage{xcolor}
\usepackage{subcaption}    
\usepackage{siunitx}
\usepackage{gensymb}
\usepackage{booktabs}

\usepackage{microtype}

\newcommand{\etal}{\textit{et al}. }

\newcommand{\projectwebsite}{https://tif-twirl-13.github.io/wholebody-pose-control}

\title{\LARGE \bf
Whole-Body End-Effector Pose Tracking
}

\author{Tifanny Portela$^{1,2}$, Andrei Cramariuc$^{1}$, Mayank Mittal$^{1,3}$ and Marco Hutter$^{1}$
\thanks{Authors are members of $^{1}$Robotic Systems Lab, ETH Zurich. $^{2}$ETH AI center. $^{3}$NVIDIA. Email:
        {\tt\small tifanny.portela@ai.ethz.ch}}%
\thanks{This project has received funding from the ETH AI Center and the European Union’s Horizon Europe Framework Programme under grant agreement No 101121321 and NCCR automation. This work has been conducted as part of ANYmal Research, a community to advance legged robotics.}%
}

\begin{document}

\maketitle
\thispagestyle{empty}
\pagestyle{empty}



\begin{abstract}
Combining manipulation with the mobility of legged robots is essential for a wide range of robotic applications. However, integrating an arm with a mobile base significantly increases the system's complexity, making precise end-effector control challenging. Existing model-based approaches are often constrained by their modeling assumptions, leading to limited robustness. Meanwhile, recent Reinforcement Learning (RL) implementations restrict the arm's workspace to be in front of the robot or track only the position to obtain decent tracking accuracy. In this work, we address these limitations by introducing a whole-body RL formulation for end-effector pose tracking in a large workspace on rough, unstructured terrains. Our proposed method involves a terrain-aware sampling strategy for the robot's initial configuration and end-effector pose commands, as well as a game-based curriculum to extend the robot's operating range. We validate our approach on the ANYmal quadrupedal robot with a six DoF robotic arm. Through our experiments, we show that the learned controller achieves precise command tracking over a large workspace and adapts across varying terrains such as stairs and slopes. On deployment, it achieves a pose-tracking error of $2.64~\si{\centi\meter}$ and $3.64 \degree$, outperforming existing competitive baselines. 
The video of our work is available at: \href{\projectwebsite}{\texttt{wholebody-pose-tracking}}.
\end{abstract}
\vspace{0.5cm}

\section{INTRODUCTION}


Over the past decade, algorithmic advancements have substantially increased legged robots' ability to traverse complex, cluttered environments and human-designed infrastructures, such as stairs and slopes~\cite{lee2020learning, Miki_2022, margolis2022walktheseways}.
Despite these improvements, their practical applicability remains constrained by their limited manipulation capabilities. Most field operations with legged robots involve minimal environmental interactions, such as visual inspections and passive load transportation.
Thus, combining a legged robot's mobility with the ability to perform manipulation tasks is critical for enhancing their applications to more real-world scenarios.

Compared to fixed base counterparts, integrating an arm onto a legged mobile platform significantly complicates the controller design because of increased degrees of freedom, redundancy and highly non-linear dynamics. To address this, the research community has mainly explored model-based and learning-based control strategies.

Model-based approaches, such as Model Predictive Control (MPC), have shown precise control on flat terrain by leveraging accurate models of the robot and its environment \cite{full_mpc_belliscoso, full_mpc_sleiman}. However, solving MPC's control problem in real-time for legged manipulators often requires the use of simplified models \cite{mpc_with_centroidal_dynamics, zmp}, which increases vulnerability to unexpected disturbances such as slipping or unplanned contacts.

In contrast, the learning-based control strategy of Reinforcement Learning (RL) has emerged as a robust alternative, directly learning control policies through interactions in simulation for locomotion~\cite{Hwangbo_2019, lee2020learning, margolis2022walktheseways} and whole-body control~\cite{fu2022deepwholebodycontrollearning}. RL enables improved resilience to external disturbances compared to MPC because of environmental variability during training~\cite{deepMimic, isaacgym, lee2020learning}. Research on legged robots, whether using legs \cite{arm2024pedipulateenablingmanipulationskills, cheng2023legsmanipulatorpushingquadrupedal} or attached arms \cite{fu2022deepwholebodycontrollearning, portela2024learningforcecontrollegged}, demonstrate that RL techniques can achieve effective end-effector position tracking across a large workspace through agile whole-body behavior. Although effective in outdoor and slippery terrains, these approaches remain unsuitable beyond flat terrain and lack orientation tracking.

\begin{figure}[t]
    \centering
    \begin{subfigure}{0.49\linewidth}
        \centering
        \begin{subfigure}{\linewidth}
            \centering
            \includegraphics[width=0.99\linewidth]{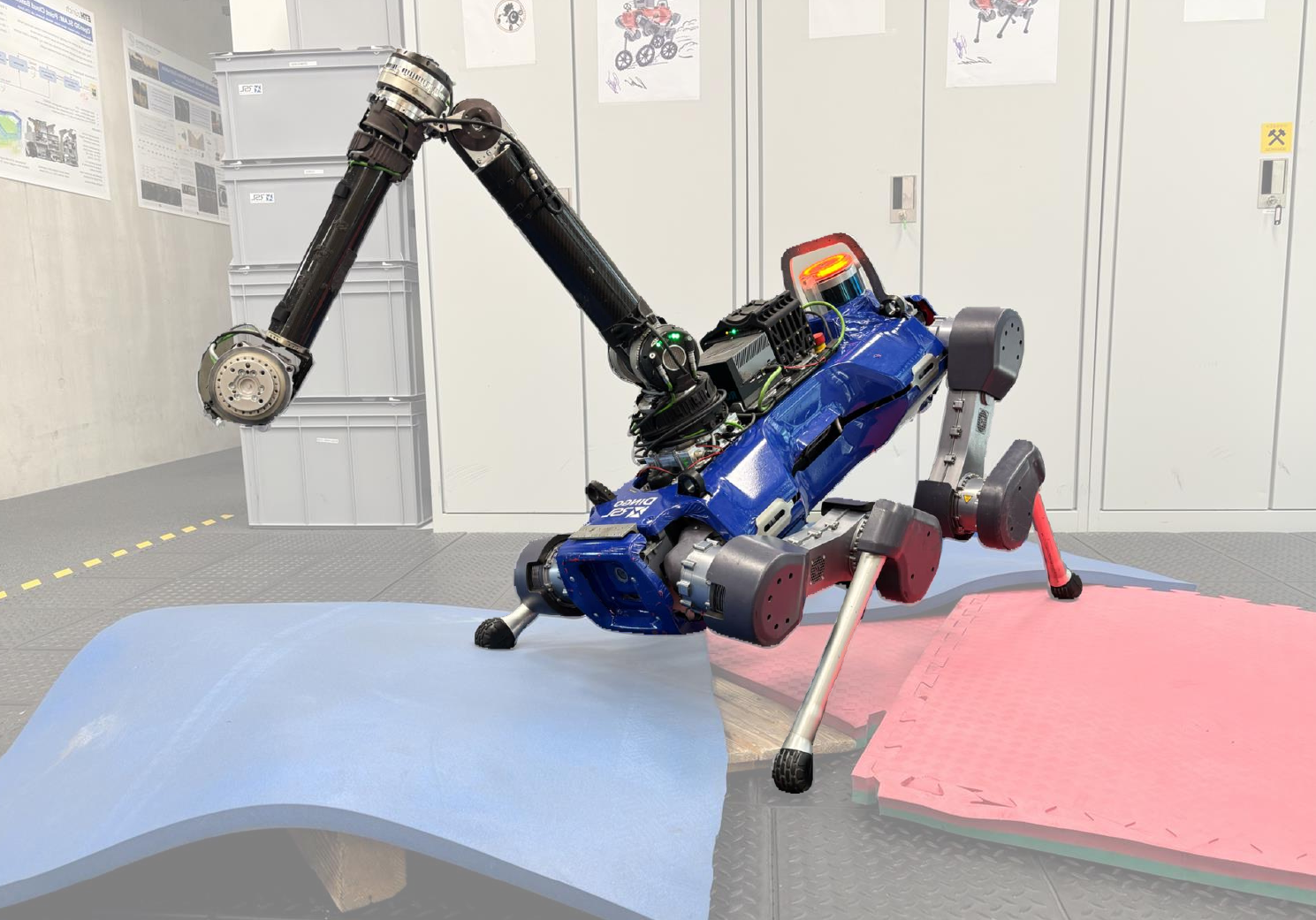}
        \end{subfigure}
        \begin{subfigure}{\linewidth}
            \centering
            \includegraphics[width=0.99\linewidth]{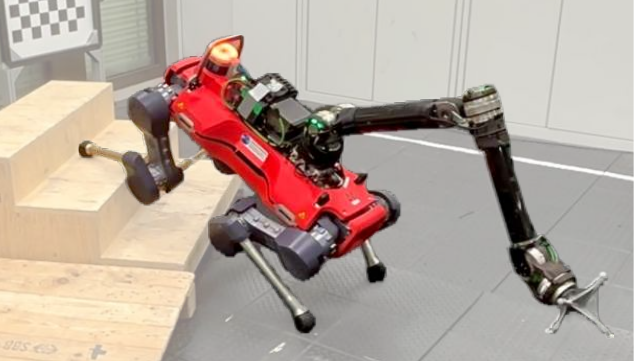}
        \end{subfigure}
        \begin{subfigure}{\linewidth}
            \centering
            \includegraphics[width=0.99\linewidth]{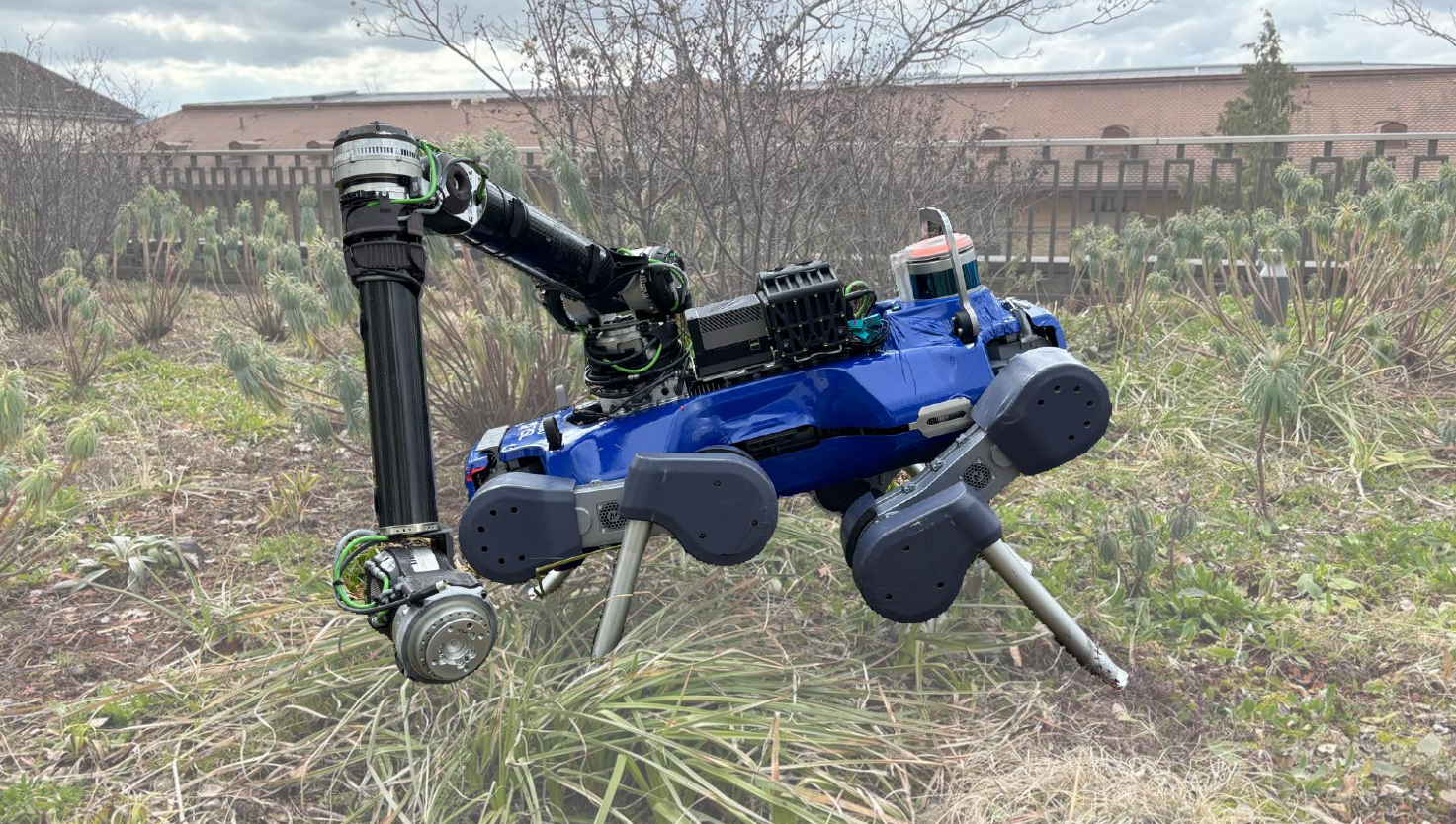}
        \end{subfigure}
    \end{subfigure}
    \hfill
    \begin{subfigure}{0.49\linewidth}
        \centering
        \includegraphics[width=\linewidth]{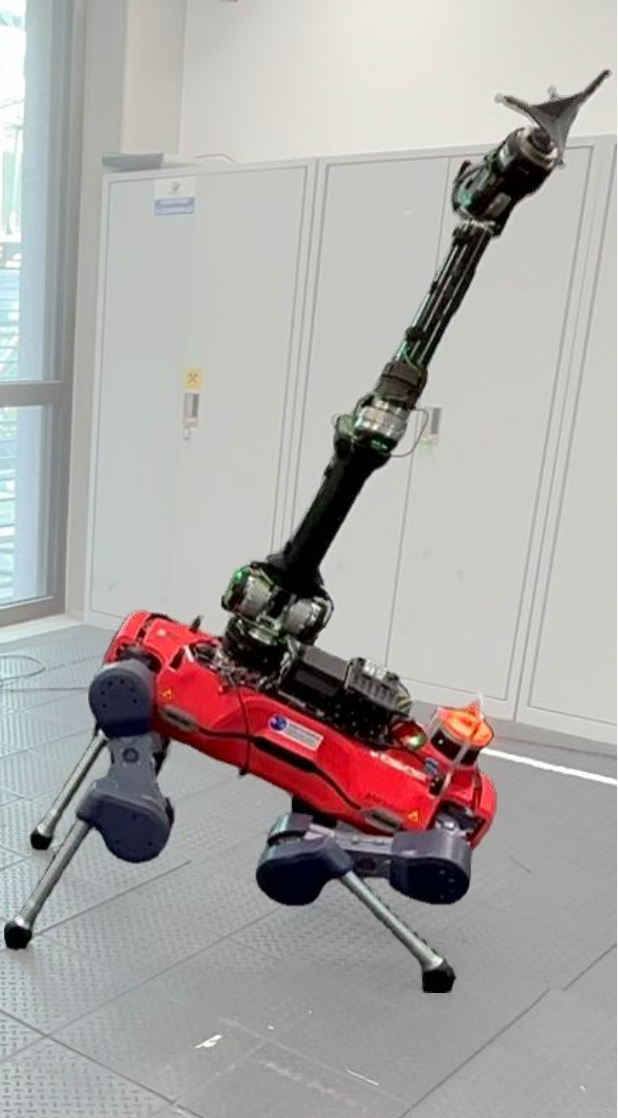}
    \end{subfigure}
    \caption{Our whole-body controller demonstrates precise end-effector pose tracking across a variety of challenging terrains, including soft mattresses, stairs and uneven natural ground.}
    \vspace{-0.6cm}
    \label{fig:hardware_wb}
\end{figure}

\begin{figure*}[ht!]
    \vspace{0.3cm}
    \centering
    \includegraphics[clip,trim=0cm 0.3cm 0cm 0.2cm,width=1.0\linewidth]{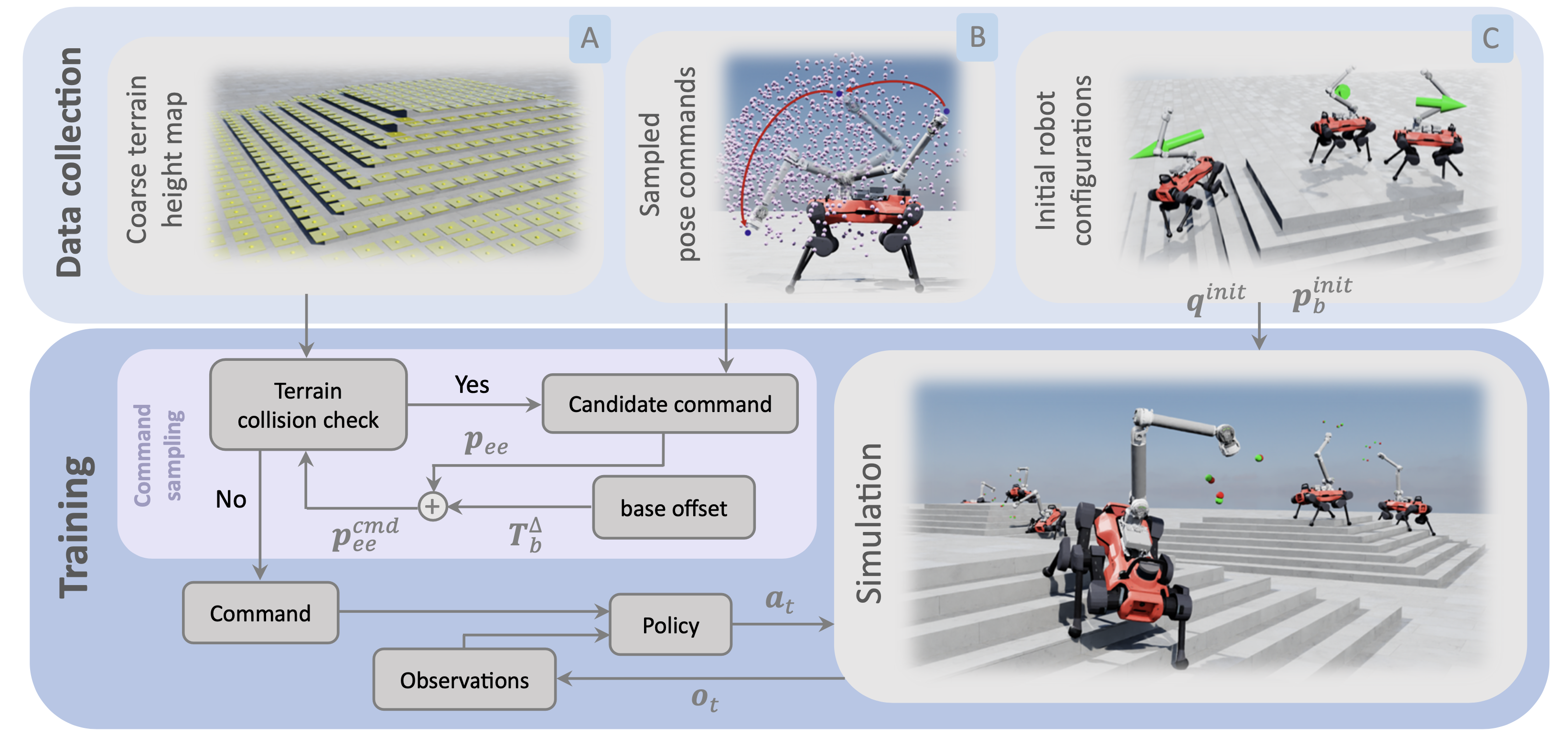}\hfill
    \caption{The training process begins with data collection, where we gather (A) the terrain mesh and a coarse terrain height map, (B) 10000 pre-sampled end-effector pose commands with a fixed base, and (C) base poses and joint angles from a pre-trained locomotion policy to initialize robots. During training, a command from (B) is slightly transformed and checked for collisions with the terrain. If collision-free, it is concatenated with observations and input to the policy; otherwise, a new command is sampled. The policy is trained in simulation with 4000 robots in parallel, outputting joint actions.}
    \label{fig:main}
\vspace{-0.4cm}
\end{figure*}

To address these shortcomings, we propose a general-purpose whole-body controller for legged robots with an attached arm. The controller is designed to provide stable end-effector pose tracking across a large operational space. Our approach includes a terrain-aware sampling strategy for end-effector pose commands, and for the robot’s initial configuration to ensure a smooth transition from a locomotion policy to the proposed whole-body controller. Experimental results show that the controller achieves precise tracking across varying terrains, such as stairs and slopes.
Our key contributions are as follows:

\begin{itemize}
    \item We propose an RL whole-body controller for 6-DoF end-effector pose tracking for quadrupeds with an arm.
    \item We showcase the learned controller's tracking capabilities over challenging terrains and its robustness when faced with external disturbances.
    \item We compare our learned controller to model-based controllers and existing RL approaches, showing higher tracking accuracy and enlarged pose reachability, both in simulation and on hardware, reaching a pose tracking accuracy of $2.64~\si{\centi\meter}$ and $3.64 \degree$.
\end{itemize}



\section{RELATED WORK}

\subsection{Model-based whole-body control}

Formulating the whole-body planning and control of legged mobile manipulators into a single optimization problem avoids treating the arm as an external disturbance to the base~\cite{hyq_with_arm,sentis2005synthesis} and helps ensure tight coordination between the base and the arm~\cite{mittal2022articulated,go_fetch}. Belliscoso~\etal~\cite{full_mpc_belliscoso} propose an online ZMP-based motion planning framework that relies on an inverse dynamics model to track generated operational-space references. In contrast, Sleiman~\etal~\cite{full_mpc_sleiman} provide a unified MPC formulation for whole-body dynamic locomotion and manipulation planning. Chiu~\etal~\cite{chiu2022collisionfreempcwholebodydynamic} further extend this formulation to account for self and environment collisions during the receding-horizon control. To increase the robustness towards external disturbances proactively, Ferrolho~\etal~\cite{ferrolho2023roloma,optimizingdynamictraj} incorporate a robustness metric into the trajectory optimization to compute offline plans for interactions under worst-case scenario. While these approaches are effective for end-effector control of legged mobile manipulators, they rely on an available system model and pre-specified contact schedule. These assumptions make model-based approaches vulnerable to large unmodeled disturbances (for instance, a heavy grasped object), slippages, and unknown terrain models.

\subsection{Learning-based whole-body control}
More recently, RL has shown to be a powerful alternative to model-based approaches for legged robots, providing robust control policies for locomotion on unstructured terrains~\cite{lee2020learning, margolis2022walktheseways} and whole-body manipulation \cite{fu2022deepwholebodycontrollearning, arm2024pedipulateenablingmanipulationskills,jeon2023learningwholebodymanipulationquadrupedal}. In the context of task-space tracking of a legged mobile manipulator's arm, existing works formulate the tracking problem using different rewards and representations for the end-effector commands.

Ma~\etal~\cite{ma2022combining} propose a hybrid approach that employs an MPC planner for the arm and RL policy for locomotion. The MPC planner outputs arm joint position and base velocity commands. Piu~\etal~\cite{roboduet} follow a similar approach but train an RL policy for the arm to replace the MPC planner. The RL policy receives the end-effector pose command as a 3D position and Euler angles. Liu~\etal~\cite{liu2024visual} leverage a hierarchical approach to train a high-level policy that provides commands for an RL policy for the base and inverse kinematics controller for the arm. However, fixing a locomotion policy for the base limits the participation of the legged base in enhancing the workspace of the arm.

Fu~\etal~\cite{fu2022deepwholebodycontrollearning} train a combined RL policy for locomotion and manipulation. For the arm, they sample position commands in spherical coordinates and orientation commands uniformly in $\mathbb{SO}(3)$. However, their results are shown only for 3D position commands and on flat terrain. A follow-up comparison in~\cite{umionlegs} reports a poor pose tracking accuracy with this work as $22.2 \si{\centi\meter}$ and $66.22 \degree$. Considering the importance of regulating forces during interactions, Portela~\etal~\cite{portela2024learningforcecontrollegged} train an RL controller for tracking end-effector 3D position and 3D force simultaneously. However, they also demonstrate their work only on flat terrains. More recently, Ha~\etal~\cite{umionlegs} use a 3D Cartesian and 6D rotation representation~\cite{zhou2020continuityrotationrepresentationsneural} for the end-effector pose and train a low-level RL policy to track a trajectory of pose commands. While they achieve a position and orientation error of $2.12 \si{\centi\meter}$ and $3.35 \degree$, respectively, the learned controller is mainly demonstrated over reference end-effector motions in front of the robot, which does not necessitate active base (leg) usage.

Our work addresses key issues in existing approaches, including poor pose tracking accuracy, limited workspace, and inability to handle complex terrains. We overcome these challenges
by developing a whole-body controller capable of precise end-effector pose tracking across a large workspace and challenging terrains, including stairs.

\section{METHOD}

We train a policy to track end-effector target poses with minimal foot displacement, intended for use alongside a locomotion policy. Figure \ref{fig:main} illustrates the overall training process. 
We use Isaac Lab~\cite{orbit} as a simulation environment to train our policy and deploy our controller on ALMA~\cite{full_mpc_belliscoso}, which integrates the Anymal D robot from ANYbotics~\cite{anybotics_anymal_specifications} with the Dynaarm from Duatic~\cite{dynaarm_specifications}. The robot has an inertial measurement unit providing body orientation and 18 actuators with joint position encoders. The main control loop and the state estimator are executed at 400 Hz while our policy runs at 50Hz on an onboard computer. While we show results on ALMA, our method remains applicable across different robot embodiments.

\begin{figure}[t!]
    \vspace{0.3cm}
    \centering
    \begin{subfigure}{0.9\linewidth}
    \includegraphics[width=0.475\linewidth]{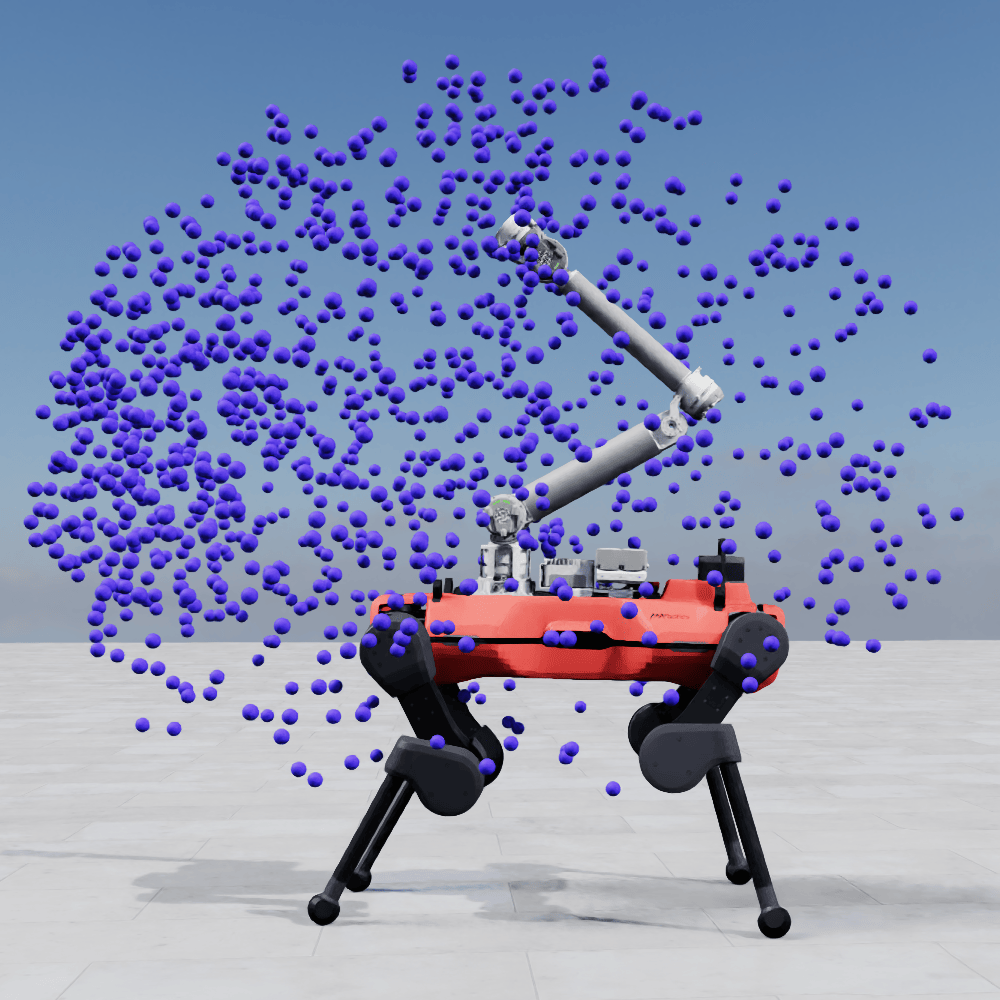}
    \hfill
    \includegraphics[width=0.475\linewidth]{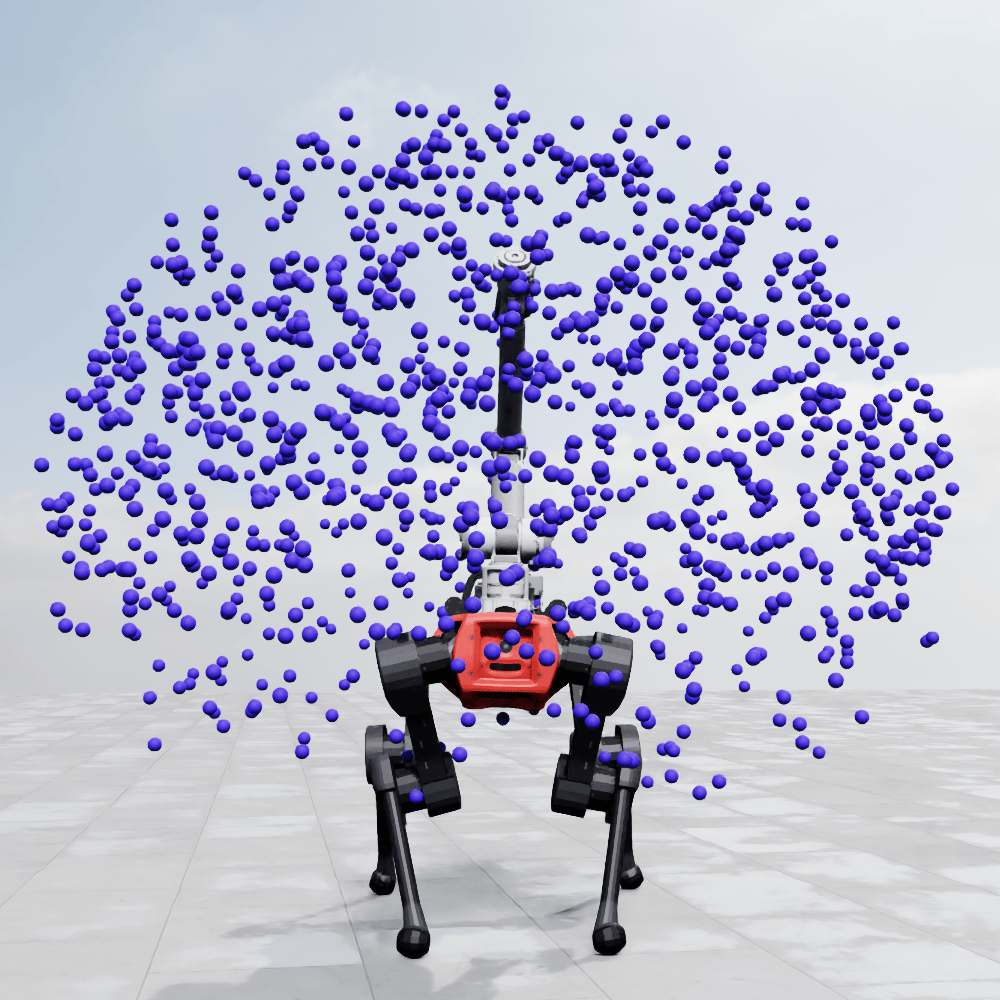}
    \caption{Initial workspace from sampling arm movements only.}
    \label{fig:workspace_views_initial}
    \end{subfigure}
    
    \vspace{0.5em} 
    
    \begin{subfigure}{0.9\linewidth}
    \includegraphics[width=0.475\linewidth]{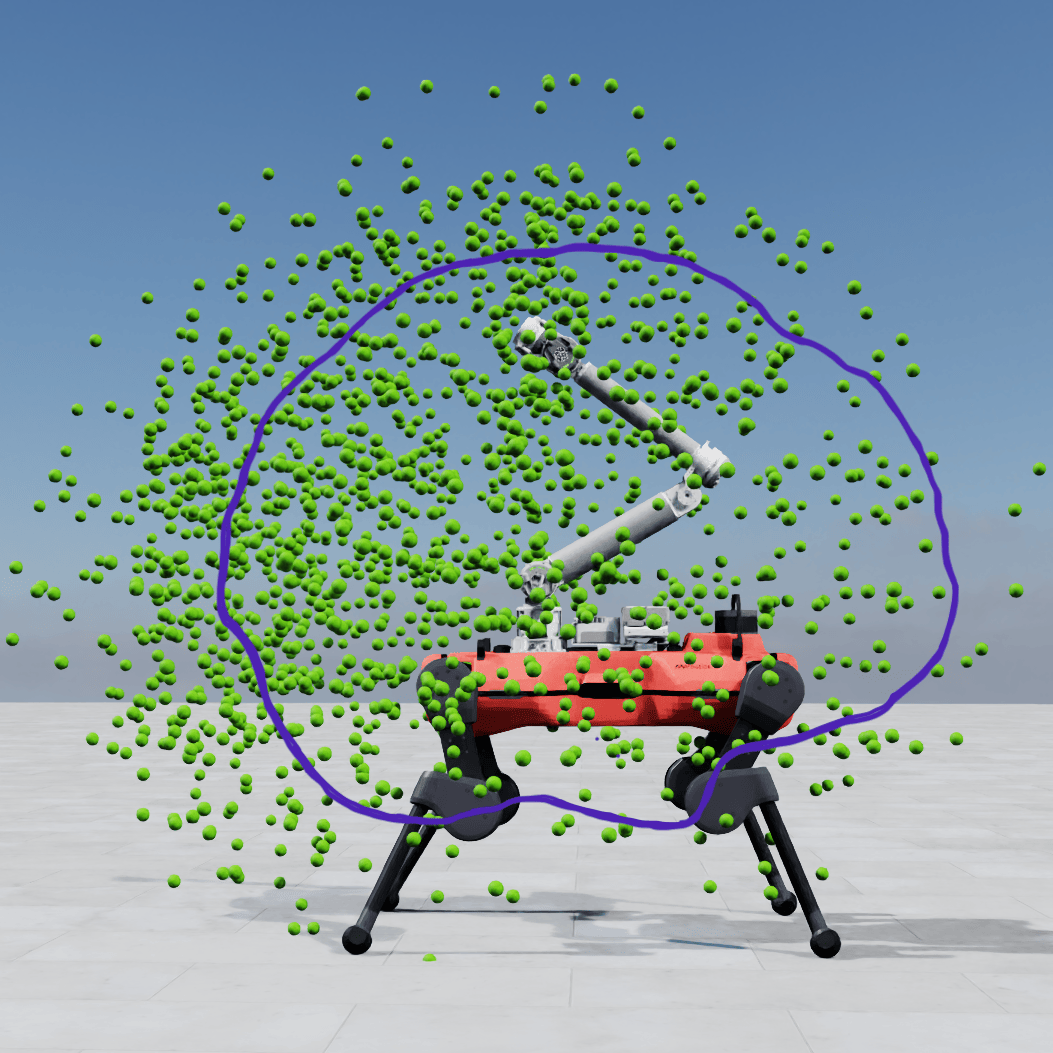}
    \hfill
    \includegraphics[width=0.475\linewidth]{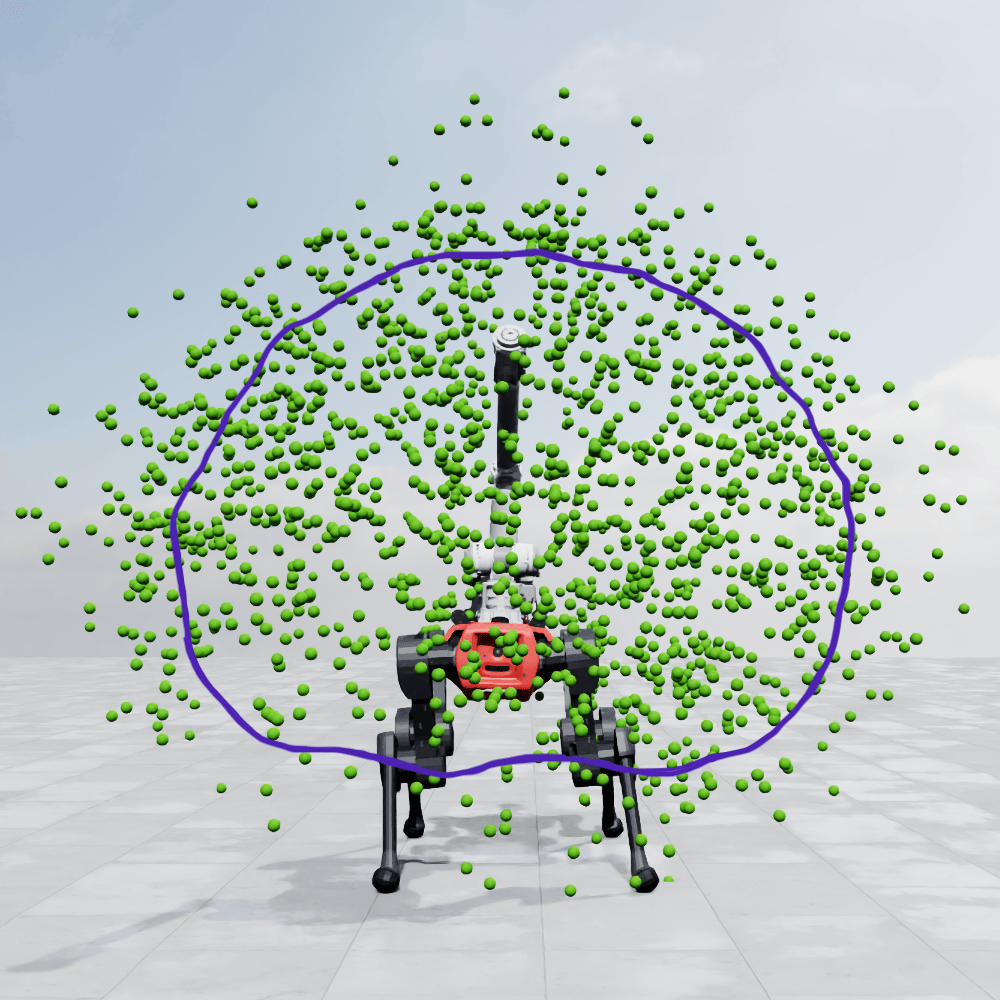}
    \caption{Expanded workspace obtained through moving the body and combining it with the initial arm-only workspace. The boundary of the initial workspace is shown in purple.}
    \label{fig:workspace_views_expanded}
    \end{subfigure}
    
    \vspace{0.5em} 
    
    \begin{subfigure}{0.9\linewidth}
    \includegraphics[width=0.475\linewidth]{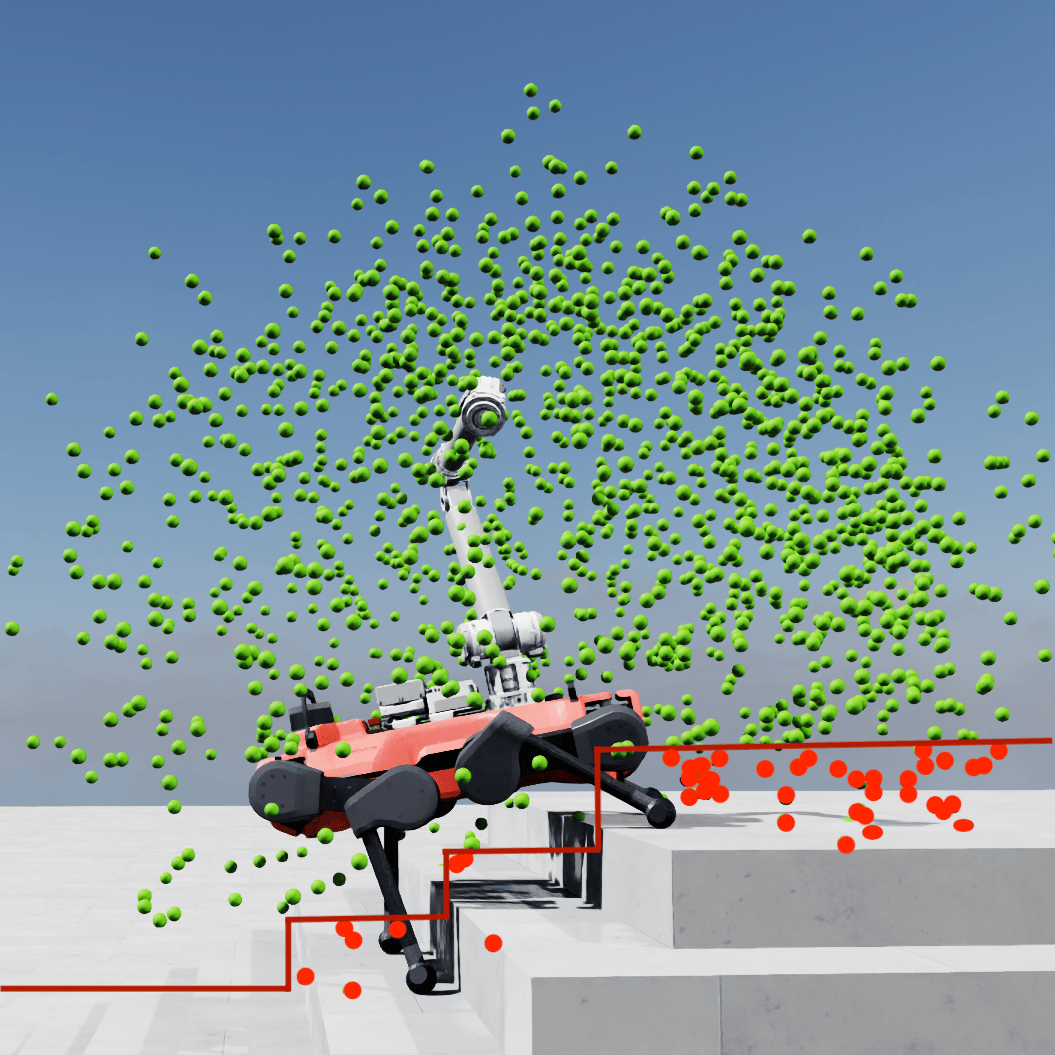}
    \hfill
    \includegraphics[width=0.475\linewidth]{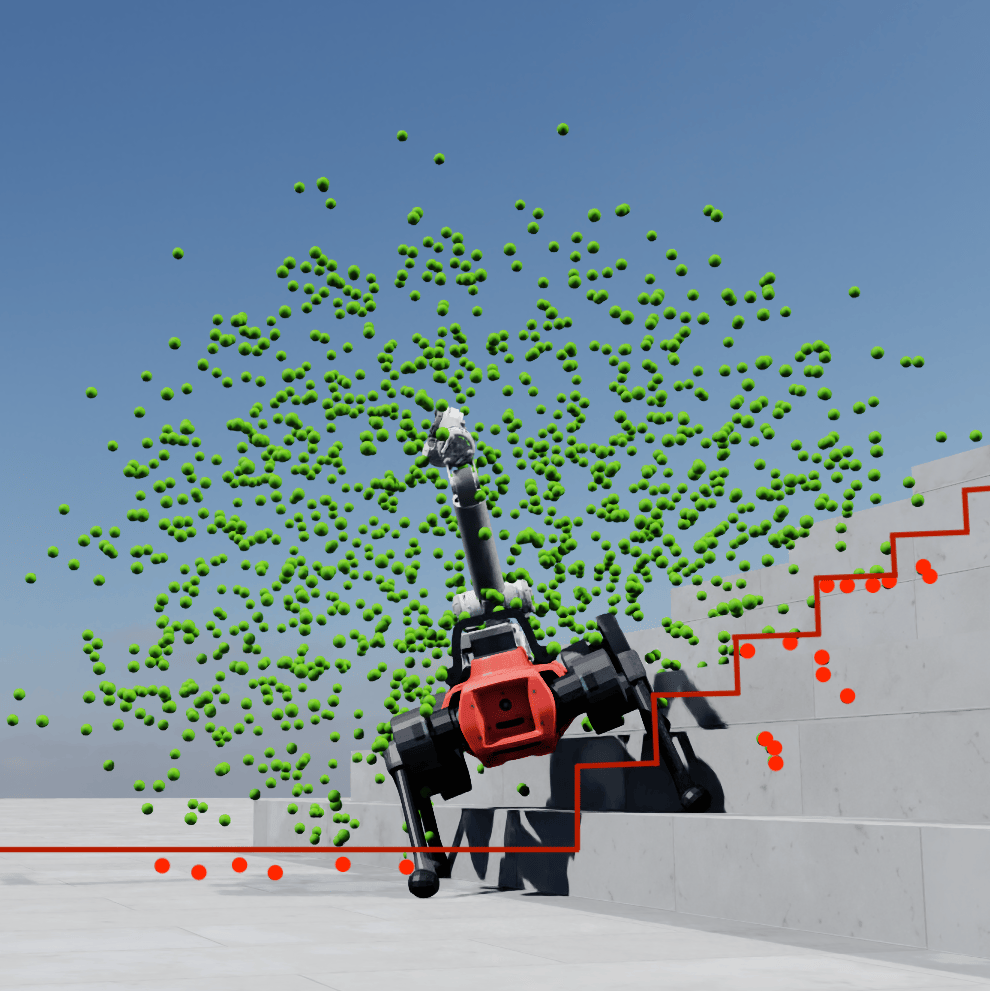}
    \caption{Expanded workspace on stairs, with poses in collision (red) removed based on the coarse terrain height map.}
    \label{fig:workspace_views_stairs}
    \end{subfigure}
    
    \caption{Front and side views of the workspace with 10000 collision-free end-effector poses (subsampled to 250 and illustrating only positions for readability).}
    \label{fig:workspace_views}
    \vspace{-0.5cm}
\end{figure}

\subsection{Policy Architecture} 
We use Proximal Policy Optimization (PPO)~\cite{ppo}, where both the actor and critic networks are implemented as multilayer perceptrons with hidden layers of size [512, 256, 128], with ELU as the activation function. The hyperparameters of the PPO algorithm are taken from prior work~\cite{rudin2022learningwalkminutesusing}.

\subsection{Command sampling} 

We pre-sample end-effector pose commands for a fixed base by iterating over the six joint angles of the arm, covering their entire range, and recording the end-effector poses that are collision-free in the base frame, as illustrated in Figure~\ref{fig:main}-B. The final dataset contains 10000 collision-free end-effector poses, with Figure~\ref{fig:workspace_views_initial} displaying a 250-pose subsample that illustrates the workspace.
If these poses were used directly as commands, a simple inverse kinematics solver for the arm would suffice. However, this command sampling strategy would limit the controller's ability to achieve certain end-effector poses that could otherwise be reached by utilizing the entire body of the mobile manipulator. To overcome this limitation, we introduce a random body pose transformation $T_{b}^\Delta \in \mathbb{R}^{6}$, applied to each pre-sampled command when a new pose is defined. This transformation is sampled within ranges of $[-0.2, 0.2]$~\si{\meter} for the x and y dimensions, $[-0.3, 0.1]$~\si{\meter} for z and, $[-\pi/6, \pi/6]$~\si{\radian} for roll, pitch and yaw. This approach ensures that the end-effector pose commands are reachable with minimal base movement. Figure \ref{fig:workspace_views_expanded} displays a 250-pose subsample of this expanded workspace.

The distribution of the pre-sampled commands is not uniform. We fit five concentric cylinders aligned along the base z-axis. The first cylinder is solid, and each subsequent cylinder has a larger radius with a hollow section. We determine their radii by first calculating the maximum radius in the $x$-$y$ plane from the 3D positions of all pre-sampled poses and dividing it into five equal parts. The distribution of poses across these cylinders, starting from the inner one, are as follows: 51\%, 21\%, 13\%, 6\%, and 2\%. To enforce more spatially uniform sampling, we randomly select one of the bins and then sample a pose from within that selected bin.

On flat terrain, this sampling procedure generally yields feasible commands (a few poses might occasionally intersect with the robot's body due to the added base offset). However, some pose commands can fall below the terrain surface on unstructured terrain like stairs, as illustrated in Figure~\ref{fig:workspace_views_stairs}. To mitigate this, we resample any command where the z-position component falls below the terrain surface height plus an 8~\si{\centi\meter} margin. We generate a coarse terrain grid map at the start of training to increase training speed and avoid real-time terrain queries. In the $x$-$y$ plane, this map stores the highest terrain point within a $20\times20$~\si{\centi\meter} patch centered on each point, as illustrated in Figure \ref{fig:main}-A. 

Finally, a new command is generated every 4 seconds, and with a 12-second episode length, the robot is exposed to 3 different commands per episode. This setup allows the robot to learn how to reach an end-effector pose from any arm configuration instead of a single-pose episode, where it would always do so from the default arm configuration.


\subsection{Command definition} 
The command of the policy is an end-effector pose $p_{ee} \in \mathbb{S}\mathbb{E}(3)$, typically represented by a separate position and orientation \cite{umionlegs, fu2022deepwholebodycontrollearning}. Separating these components introduces two main challenges. First, defining a rotation representation that is easily learnable is difficult~\cite{jonas_rotation}. Second, this separation requires a fixed trade-off between position and orientation rewards, which may not be optimal for all workspace poses.

To avoid these issues, we use a keypoint-based representation similar to that used in~\cite{allshire_2022} for in-hand cube reorientation, which has been shown to improve the ability of the RL algorithm to learn the task at hand. In this formulation, the keypoints represent the vertices of a cube centered on the end-effector's position and aligned with its orientation. While 8 corner points fully define the cube, we use the minimum required of 3 keypoints with direct correspondence between measured and target poses to uniquely and completely define the pose. The side length of the cube is set to 0.3 meters.

\subsection{Action and Observation Space}
The robot's movement is managed through an eighteen-dimensional space ($a^t \in \mathbb{R}^{18}$). This action space controls position targets for a proportional-derivative controller applied to each robot joint. The joints include the legs' thigh, calf, and hip joints and the six arm joints. The position targets are derived as $\sigma_a a^t + q_{def}$, where $\sigma_a = 0.5$ is a scaling factor, and $q_{def}$ represents the robot's default joint configuration, which corresponds to the robot standing with its arm raised.

The observation, represented as $o^t$, relies solely on proprioceptive information. Its elements consist of the gravity vector projected in the base frame $g_{b}^t \in \mathbb{R}^{3}$, the base linear and angular velocities, $v_{b}^t \in \mathbb{R}^{6}$, the joint positions, $q^t \in \mathbb{R}^{18}$ and the previous actions $a^{t-1} \in \mathbb{R}^{18}$:
\begin{equation}
    o^t = [g_{b}^t, v_{b}^t, q^t, a^{t-1}] \in \mathbb{R}^{45} 
\end{equation}

The observation, $o^t$, is augmented with the end-effector pose command for the policy input. This command is defined as the positional difference between the current and the target keypoints of the end-effector in the base frame $\prescript{b}{}{\!p_{ee}^{\text{cmd}}} \in \mathbb{R}^{9}$, where each of the three keypoints provides a 3D position vector, resulting in a 9-dimensional vector in total.
The observation does not include explicit terrain information. Nevertheless, as the robot poses are sampled from collision-free configurations (\ref{sec:init_poses}) and the robot remains mostly stationary, it can deduce terrain properties indirectly.

\subsection{Rewards}
The final reward \( R \) is the sum of the task rewards \( R_T \) and penalties \( R_P \): \( R = R_T + R_P \). The task reward \( R_T \) can be divided into four subcategories: tracking, progress, feet contact force, and initial leg joint rewards: \( R_T = \omega_1 R_{t} + \omega_2 R_{p} + \omega_3 R_{f} + \omega_4 R_{q} \), where \( \omega_1 = 13 \), \( \omega_2 = 80 \), \( \omega_3 = 0.015 \) and \( \omega_4 = 0.4 \).

\textbf{Tracking Reward (\( R_{t} \))} is a delayed reward focused on tracking the three keypoints, representing the end-effector pose command, during the last two seconds (\( T_r = 2\si{\second}\)) of a 4-second command cycle (\( T = 4\si{\second}\)). Delaying the reward emphasizes the importance of being in the correct pose during the final 2 seconds without penalizing the path to getting there. This prevents the unwanted behavior that continuous rewards might encourage, such as passing through the robot's body when transitioning from a pose on one side of the robot to a target on the opposite side.

\[
R_{t} =
\begin{cases} 
    \frac{1}{T_r} \sum_{k=0}^{3} e^{\frac{1}{\sigma_t}\lVert \prescript{b}{}{\!p_{ee, k}^{\text{meas}}} -  \prescript{b}{}{\!p_{ee, k}^{\text{cmd}}} \rVert_2} & \text{if } t > T - T_r \\
    0 & \text{otherwise}
\end{cases}
\]

Here, \( \prescript{b}{}{\!p_{ee, k}^{\text{meas}}} \) and \( \prescript{b}{}{\!p_{ee, k}^{\text{cmd}}} \) are the positions of the measured and commanded keypoints in \( \mathbb{R}^3 \), respectively, and \( \sigma_t = 0.05 \).

\textbf{Progress Reward (\( R_p \))} addresses the sparsity of the tracking reward and reduces end-effector oscillations by incentivizing steady progress toward the target. It compares the current distance \( d^t \in \mathbb{R}^3 \) between the measured and commanded keypoints to the smallest previously recorded distance \( d  \in \mathbb{R}^3 \). If \( d^t \) is smaller, the reward is calculated as:

\[
R_{p} =
\begin{cases} 
    \frac{1}{3} \sum_{k=0}^{3} (d_k - d^t_k) & \text{if } d^t < d \\
    0 & \text{otherwise}
\end{cases}
\]


\textbf{Feet Contact Force Reward (\( R_f \))} encourages the robot to maintain ground contact with all four feet. The reward is non-zero only if all four feet are firmly in contact with the ground. To account for small disturbances and ensure genuine contact, 1 Newton is subtracted from the force on each foot, denoted as $F_i$. The reward is calculated as the sum of these adjusted forces:

\[
R_{f} =
    \sum_{i=1}^{4} \text{max}(F_i - 1, 0)
\]

\textbf{Initial Leg Joint Reward (\( R_q \))} encourages the robot to maintain its leg joints in a configuration close to those sampled from the locomotion policy, as these are known to result in a stable posture, with \( \sigma_q\) defined as \( 0.05 \).
\[
R_{q} =  \sum_{i=0}^{12} e^{\frac{1}{\sigma_q}(q_i^{\text{init}} - q_i)}
\]

\textbf{Penalties (\( R_P \))} penalize joint torques, joint accelerations, action rates and target joint positions above the limits: \( R_P = \omega_5 \lVert \boldsymbol{\tau} \rVert^2 + \omega_6 \lVert \overset{\text{\tiny $\cdot$}\text{\tiny $\cdot$}}{\boldsymbol{q}} \rVert^2 + \omega_7  \lVert \boldsymbol{a}_t -  \boldsymbol{a}_{t-1} \rVert^2 + \omega_8  \lVert \boldsymbol{q} - \boldsymbol{q}_{\text{lim}} \rVert_1\), where \( \omega_5 = -3e^{-5} \), \( \omega_6 = -3e^{-6} \), \( \omega_7 = -5e^{-2} \) and \( \omega_8 = -1.3\). Finally, we terminate on knee and base contacts. 

\subsection{Terrains and Curriculum training}
The robots are trained in simulation on four procedurally generated terrains: flat, randomly rough, discrete obstacles, and stairs, as defined in~\cite{rudin2022learningwalkminutesusing}. Gradually increasing task difficulty during training has been shown to enhance learning
\cite{lee2020learning,xie2020allstepscurriculumdrivenlearningstepping, rudin2022learningwalkminutesusing}. We employ a terrain curriculum similar to the one proposed in \cite{rudin2022learningwalkminutesusing}, adapted for end-effector pose tracking. If the average position tracking error across the three commands within an episode is under $20$~\si{\centi\meter} when the task reward is active (during the final 2 seconds of each 4-second command), and the average orientation tracking error is under $20\degree$, the terrain difficulty increases after the next reset. The terrain difficulty decreases if the error exceeds $80$~\si{\centi\meter} and $120\degree$. Robots at the highest level are assigned a random level to prevent catastrophic forgetting.

\subsection{Initial poses}
\label{sec:init_poses}
The whole-body end-effector pose tracking policy proposed in this work does not include locomotion capabilities. This decision is motivated by the observation that very few tasks require simultaneous locomotion and active object manipulation. Therefore, this policy is designed to operate alongside a separate locomotion policy. 

To facilitate the learning of a stable leg posture, the initial base pose $p_{b}^{init} \in \mathbb{S}\mathbb{E}(3)$ and joint angles $q^{init} \in \mathbb{R}^{18}$ of the robots upon reset are taken from a pre-trained locomotion policy~\cite{Arm_2023}. This policy, designed for the same mobile-legged manipulator using RL, incorporates arm joint positions and velocity measurements in its observation space. It only controls the leg joints, enabling rough terrain locomotion with any arm configuration. This locomotion policy takes a three-dimensional base velocity command as input: linear velocities along the x and y axes and yaw rotation.

Before training, robots are spawned at the center of the terrain and given randomized heading commands in the interval \([-1, 1]\) \(\si{\radian\per\second}
\) and linear velocity commands in the interval \([-1, 1]\) \(\si{\meter\per\second}
\) for a 4-second period, as illustrated in Figure \ref{fig:main}-C.
After this time, the terrain, the base pose $p_{b}$, and the joint angles $q$ are recorded for robots that remain in stable configurations, defined as those with an angle difference between the gravity vector and their base-aligned projected gravity vector less than $55\degree$. When training begins, the saved terrain and robot configurations are loaded. This process also ensures a smooth transition from the locomotion to the whole-body policy, as omitting this step causes the robot to jump when switching between policies.

\begin{figure}[t!]
    \centering
    \vspace{0.2cm}
    \includegraphics[width=8.5cm]{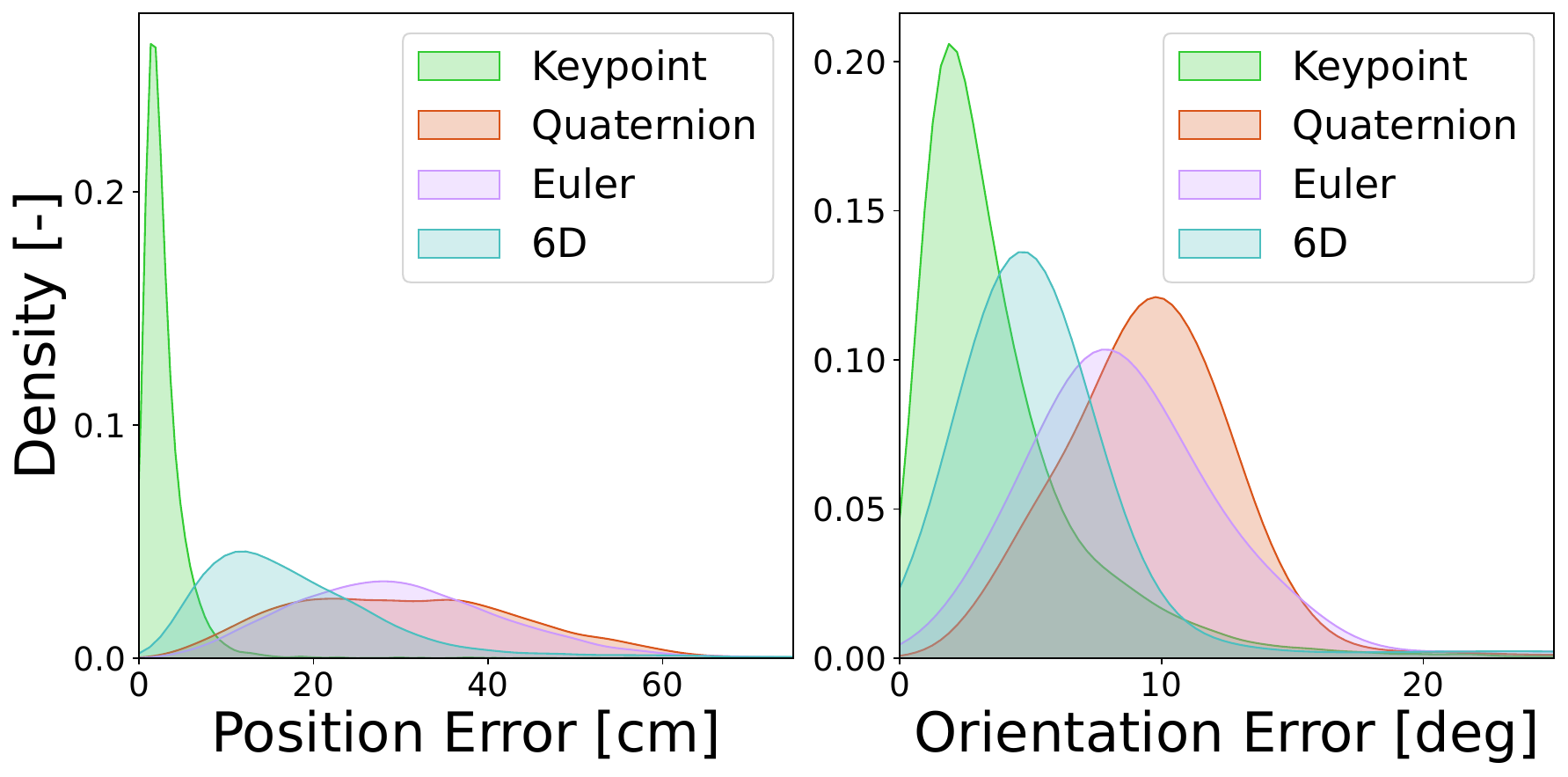}
    \caption{Distribution of the position and orientation errors for 10000 end-effector pose commands measured on flat terrain in simulation for four different pose representations.}
    \label{fig:tracking_comparison_sim}
    \vspace{-0.5cm}
\end{figure}

\subsection{Sim-to-Real}
When the training starts, we add a mass on the end-effector randomly sampled from the interval \([0, 2.0]\)~\si{\kilo\gram}, and the inertia of this rigid body links is scaled by the ratio between the new mass and the original one.
Random perturbances are applied to the end-effector, such as an impulse force sampled from the interval \([-10, 10]\)~\si{\newton} every 3 to 4 seconds, and random pushes on the robot's base simulated as base velocity impulses sampled from the interval \([-0.5, 0.5]\)~\si{\meter\per\second} along the x-y dimensions. Random noise is also added to the observations.

\section{RESULTS AND DISCUSSION}

\subsection{Simulation experiments}
\subsubsection{Pose representation comparison}
To analyze different representations for end-effector pose commands, we evaluate the tracking performance of our keypoint-based pose representation against three other representations from related works: a 3D vector for end-effector position combined with (A) a quaternion \cite{fu2022deepwholebodycontrollearning}, (B) with Euler angles \cite{roboduet}, or (C) with a 6D vector representation \cite{umionlegs}. 
For each representation, the command is adapted to include both the position and orientation differences expressed in the base frame. Specifically, for (A), we use quaternion multiplication with the inverse of the measured quaternion, we use the difference of Euler angles for (B), and the difference between 6D vectors for (C).

Our tracking reward is adapted to $R_{t} = \frac{1}{T_r} e^{\frac{1}{\sigma_t}(\Delta_{\text{pos}}^t + \Delta_{\text{rot}}^t)}$ for $t > T - T_r$, where $\sigma_t = 0.15$. The progress reward is given by $R_{p} = (\Delta_{\text{pos}} - \Delta_{\text{pos}}^t) + (\Delta_{\text{rot}} - \Delta_{\text{rot}}^t)$, provided that both $\Delta_{\text{pos}}^t < \Delta_{\text{pos}}$ and $\Delta_{\text{rot}}^t < \Delta_{\text{rot}}$. The position error $\Delta_{\text{pos}}^t$ is calculated as the norm of the difference in 3D position, while the orientation error $\Delta_{\text{rot}}^t$ is derived from the rotational difference between quaternions. For all pose representations, including (A), (B), and (C), we convert the rotations back to quaternions, multiply one quaternion by the conjugate of the other, and then convert the result to an axis-angle representation to calculate the error.

As shown in Figure \ref{fig:tracking_comparison_sim}, the keypoint-based pose representation significantly outperforms the other three, with the 6D representation (C) ranking second with an average tracking error of \( 16.03\)~\si{\centi\meter} and \( 3.87 \degree\) larger than ours. Both quaternion (A) and Euler angles (B) representations yield similar results, with average errors \( 27\)~\si{\centi\meter} and \(6.3 \degree\) exceeding our approach. The superior performance of the keypoint-based and 6D representations, compared to the discontinuous quaternion and Euler angle representations, likely stems from their continuous nature, as discussed in \cite{jonas_rotation}. Tuning rewards for pose tracking with separate terms for position and orientation proved challenging, due to the difficulty of balancing two quantities with different units and magnitudes. Achieving both position and orientation tracking in (A), (B), and (C) required many iterations, and often, the training would collapse, prioritizing either position or orientation tracking but rarely both. In contrast, the keypoint-based pose representation required far less tuning due to its unified representation of both aspects.

Table \ref{tab:mass_errors} presents the average position and orientation errors on flat terrain for different added masses on the end-effector in simulation. Within the training range $[0, 2.0]$~\si{\kilo\gram}, the tracking errors remain stable, with an average position error of \(0.83\)~\si{\centi\meter} and orientation error of \(3.45 \degree\). When the added mass exceeds the training distribution, the tracking performance degrades with errors reaching \(15.33\)~\si{\centi\meter} and \(45.02 \degree\) for a \(4.5\)~\si{\kilo\gram} load. This highlights the controller's robustness beyond the training range while also demonstrating its limitations when encountering significantly higher payloads.

\subsubsection{Comparison to model-based control}
We compare the tracking performance of our whole-body RL policy with the model-based whole-body MPC controller from prior work \cite{chiu2022collisionfreempcwholebodydynamic} optimized for the same robot.
For this comparison, we fine-tuned the original weight parameters to optimize whole-body behavior. To enable more natural movement, we set the MPC base reference weights to zero. We evaluate both controllers on flat terrain using the same 35 end-effector pose commands in the expanded workspace (Figure \ref{fig:workspace_views_expanded}). Both controllers perform similarly in terms of median errors, with the RL controller achieving \(1.81\)~\si{\centi\meter} / \(1.73 \degree\), and the MPC controller \(2.17\)~\si{\centi\meter} / \(1.53 \degree\). However, the mean errors are notably higher for the MPC controller, reaching \(6.43\)~\si{\centi\meter} / \(6.88 \degree\), while the RL controller maintains significantly lower values at \(2.21\)~\si{\centi\meter} / \(2.01 \degree\). This discrepancy results from the MPC controller's inability to manage the trade-off between pose tracking and self-collision avoidance, causing the arm to get stuck near the base during transitions between distant poses -- an issue that our RL controller effectively avoids.

Since the MPC controller was tuned specifically for flat terrain, we did not compare its performance with our whole-body RL policy on stairs. Unlike our RL policy, the MPC controller would require a terrain model to handle rough terrain, which was beyond the scope of our evaluation.

\subsection{Hardware experiments}

\subsubsection{Pose tracking accuracy}
\begin{figure}[t!]
\centering
\vspace{0.2cm}
\includegraphics[width=8.5cm]{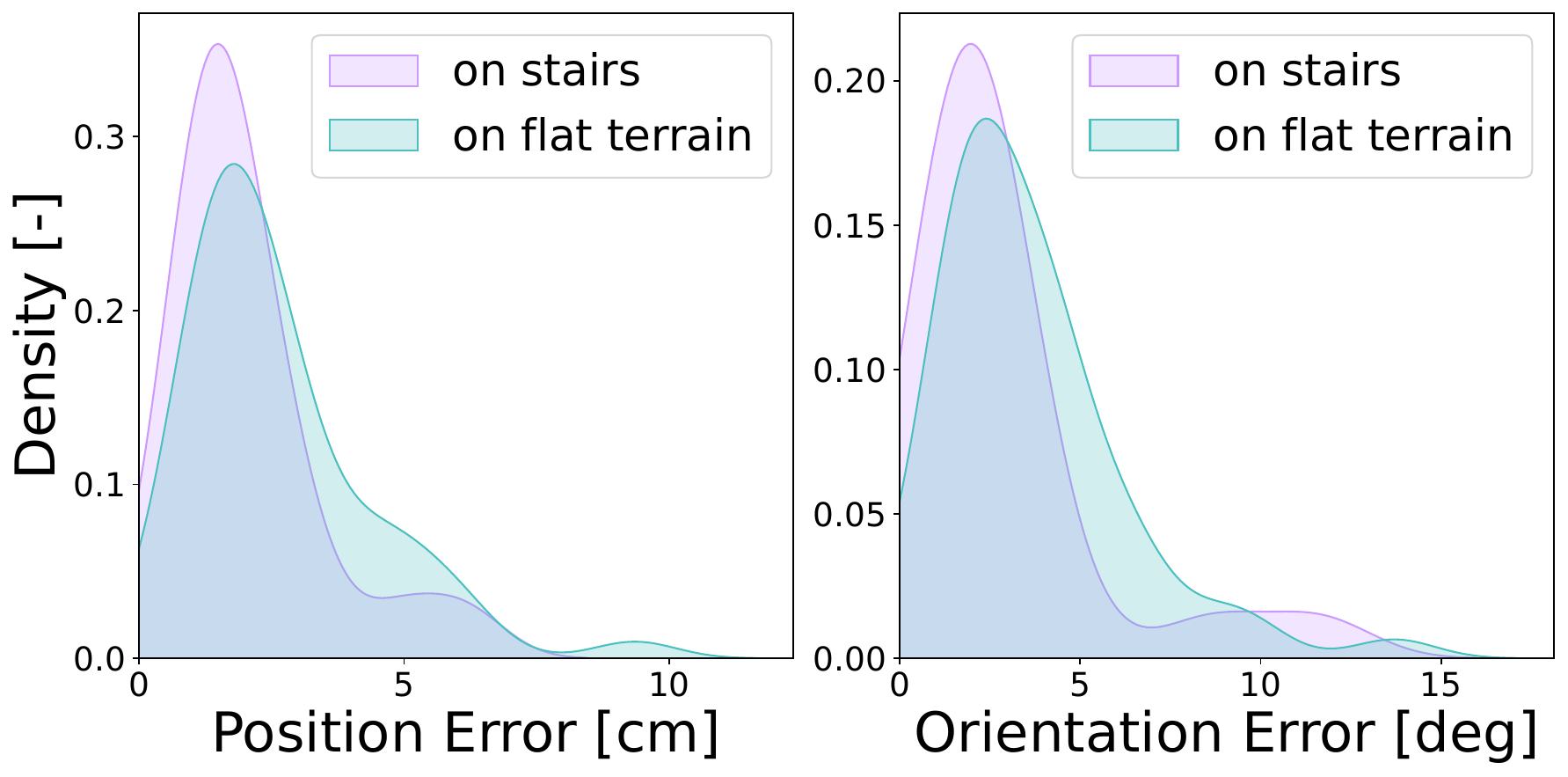}
\caption{Distribution of the position and orientation errors for 20 end-effector pose commands, measured on hardware, on both flat terrain and stairs.}
\label{fig:tracking_hardware}
\vspace{-0.5cm}
\end{figure}

We assess our controller's tracking performance using a motion capture system across 20 randomly sampled poses in the expanded workspace (Figure \ref{fig:workspace_views_expanded}). Poses are sent sequentially, with substantial changes in position and orientation, resulting in effective whole-body behavior as shown in Figure \ref{fig:hardware_wb}. The average error reaches \(2.03\)~\si{\centi\meter} and \(2.86 \degree\). These results, which are illustrated in Figure \ref{fig:tracking_hardware}, closely match the performance observed in simulation, demonstrating a minimal sim-to-real gap.

\begin{table}[t]
    \centering
    \caption{Average position $\bar{e}_p$ and orientation $\bar{e}_o$ errors for different added masses on the end-effector $m_a$ for 10000 end-effector pose commands on flat terrain in simulation.}
    \begin{tabular}{c|ccccccc}
        \toprule
        \text{$m_a$ [kg]} & [0 - 2.0] & 2.5 & 3.0 & 3.5 & 4.0 & 4.5 \\
        \midrule
        \text{$\bar{e}_p$ [cm]} & 0.83 & 1.18 & 1.89 & 4.77 & 10.69 & 15.33 \\
        \text{$\bar{e}_o$ [deg]} & 3.45 & 6.99 & 10.87 & 22.54 & 36.31 & 45.02 \\
        \bottomrule
    \end{tabular}
    \label{tab:mass_errors}
\end{table}

\subsubsection{Robustness to external disturbances}

We evaluate the tracking performance of our whole-body RL policy on stairs using a motion capture system across 20 sampled poses in the expanded workspace (Figure \ref{fig:workspace_views_expanded}) in the half-space in front of the robot under three base orientations: sideways on the stairs, facing up and facing down, as shown in Figure \ref{fig:hardware_wb}. The average position error reaches \(2.64\)~\si{\centi\meter}, and the average orientation error \(3.64 \degree\). Figure \ref{fig:tracking_hardware} shows that the tracking performance remains consistent with that on flat terrain.

When switching orientations on the stairs a locomotion policy is used~\cite{Arm_2023} and the transition between policies is smooth thanks to the robot initialization process described in Section \ref{sec:init_poses}. Without this initialization step, the robot experiences jumps when transitioning between policies.

Additionally, the system can handle up to \(3.75\) \(\text{kg}\) of weight on the end-effector when stationary, and up to \(1.3\) \(\text{kg}\) in movement. This flexibility is advantageous as it avoids the need to model weight changes, typically required in model-based approaches, allowing for the attachment of various end-effectors and dynamic carrying of unknown payloads during operation.

\section{CONCLUSION}
We have presented a whole-body RL-based controller for a quadruped with an arm that can reach even the most difficult poses. Our controller achieves high accuracy also on rough terrain (\textit{e.g.}, on stairs), which we show in real-world experiments with ANYmal with an arm. Additionally, our contributions include providing a formulation for learning pose tracking that is superior to existing methods with poor accuracy or only considering position tracking. Our hardware experiments show an average tracking accuracy of $2.64~\si{\centi\meter}$ for position and $3.64 \degree$ for orientation on challenging terrain.
Future work includes integrating 3D environment representations for self-learned collision avoidance, as in~\cite{miki2024learningwalkconfinedspaces}, and improving tracking performance under heavy, unmodeled payloads by incorporating a memory network (e.g., LSTM~\cite{hochreiter1997long}) or a concurrent state estimation architecture~\cite{Ji_2022}.




\addtolength{\textheight}{-2.0cm}   

\bibliographystyle{IEEEtran}
\bibliography{IEEEabrv,IEEEfull}{}

\end{document}